\pdfoutput=1

\documentclass[11pt]{article}

\usepackage{emnlp2021}

\usepackage{times}
\usepackage{latexsym}
\usepackage{graphicx}
\usepackage{booktabs}
\usepackage{mdframed}
\usepackage{tikz}

\usepackage[T1]{fontenc}

\usepackage[utf8]{inputenc}

\usepackage{microtype}

\title{Open-Domain Question-Answering for COVID-19 \\and Other Emergent Domains}

\author{Sharon Levy\S, Kevin Mo\P, Wenhan Xiong\S, William Yang Wang\S \\
  \S University of California, Santa Barbara \\
  \P Princeton University \\
  \texttt{\{sharonlevy,xwhan,william\}@cs.ucsb.edu}, \texttt{kevinmo@princeton.edu} \\}

\begin{document}
\maketitle
\begin{abstract}
Since late 2019, COVID-19 has quickly emerged as the newest biomedical domain, resulting in a surge of new information. As with other emergent domains, the discussion surrounding the topic has been rapidly changing, leading to the spread of misinformation. This has created the need for a public space for users to ask questions and receive credible, scientific answers. To fulfill this need, we turn to the task of open-domain question-answering, which we can use to efficiently find answers to free-text questions from a large set of documents. In this work, we present such a system for the emergent domain of COVID-19. Despite the small data size available, we are able to successfully train the system to retrieve answers from a large-scale corpus of published COVID-19 scientific papers. Furthermore, we incorporate effective re-ranking and question-answering techniques, such as document diversity and multiple answer spans. Our open-domain question-answering system can further act as a model for the quick development of similar systems that can be adapted and modified for other developing emergent domains.

\end{abstract}

\section{Introduction}\label{sec:intro}

With the rise of social media and other online sources, it is easy to access information from sites without third-party filtering ~\cite{allcott2017social}. As such, it is important in today’s society to create systems that can provide credible and reliable information to users. This is especially true in the context of emergent domains which, unlike more established sectors, may contain rapidly changing information. COVID-19 follows this pattern, with over 100,000 related articles published in 2020 and new research findings still frequently reported \cite{else2020torrent}. 

However, the vast interest and exposure surrounding this topic have consequently generated a rise in misinformation \cite{kouzy2020coronavirus, medina-serrano-etal-2020-nlp}. This can lead to lower compliance with various preventative measures such as social distancing, which in turn can continue the spread of the virus \cite{bridgman2020causes,tasnim2020impact}. A question-answering system that allows users to ask free-text questions with answers deriving from published articles and reliable scientific sources can help mitigate this spread of misinformation and inform the public at the same time.

The task of open-domain question-answering has risen in prominence in recent years \cite{chen-etal-2017-reading,yang2019end,xiong2020answering}. Systems have evolved from keyword-based approaches~\cite{salton1986introduction} to the utilization of neural networks with dense passage retrieval~\cite{xiong-etal-2021-progressively}. Furthermore, large-scale datasets have been used to train and test these systems, such as general knowledge datasets \cite{joshi2017triviaqa,nguyen2016ms} and domain-specific datasets\footnote{\url{https://trec.nist.gov/data.html}} \cite{tsatsaronis2012bioasq}. However, many of these systems are evaluated on these established datasets with abundant questions and clearly defined answers. In the case of an emergent domain system, this likely will not be available and the reduced data size can result in lower answer precision. 

\begin{figure*}[t]
  \centering
  \includegraphics[width=\linewidth]{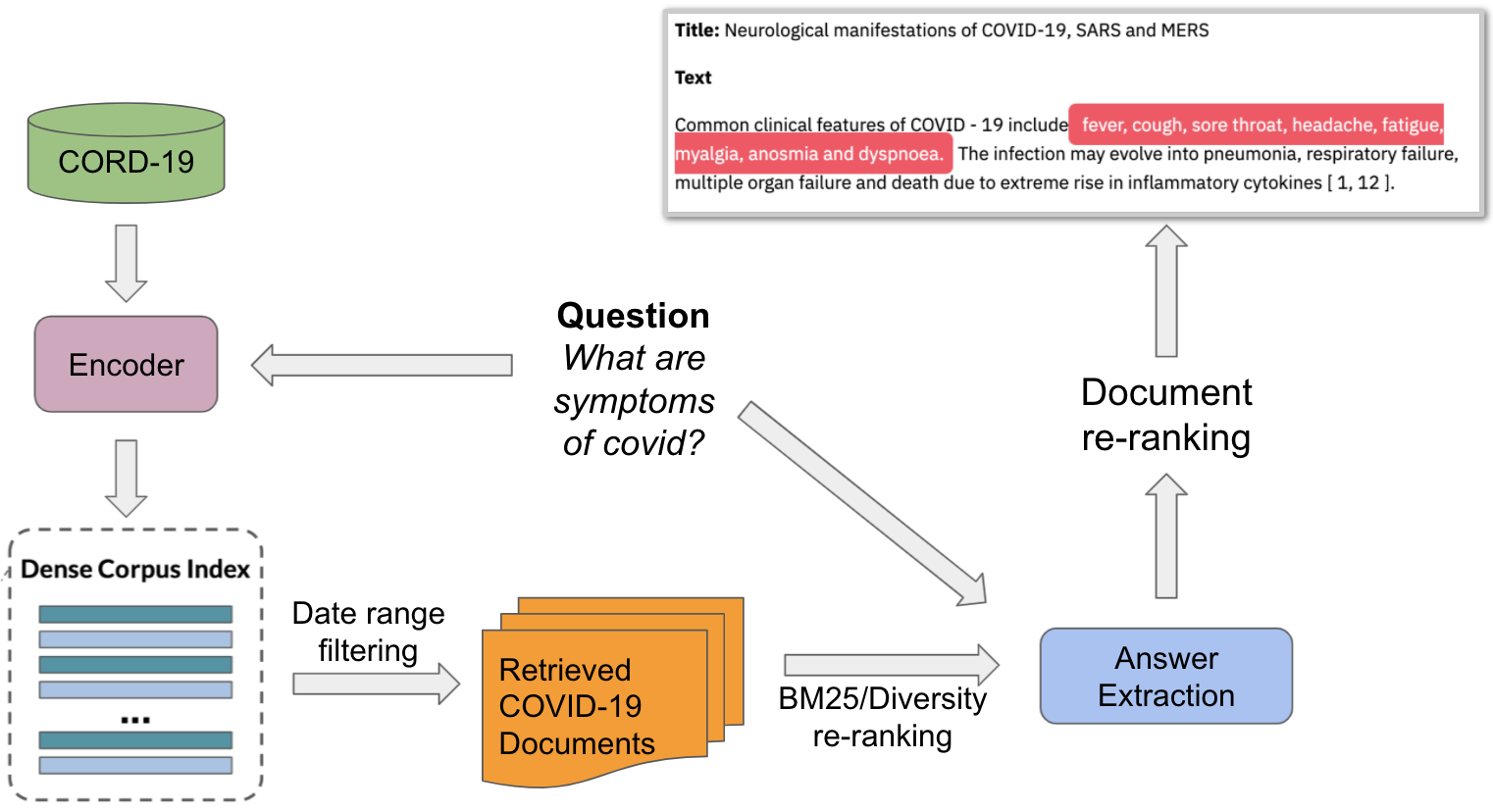}
  \caption{An overview of the COVID-19 open-domain question-answering system. The retrieval component is shown on the left and the reading comprehension/answer extraction component is shown on the right.}\label{fig:overall}
\end{figure*}

 In this paper, we build an open-domain question-answering system in the emergent domain of COVID-19. We aim to overcome a staple issue with emergent domain question-answering systems: lack of data. While several COVID-19-related datasets have been published since the beginning of the pandemic \cite{trec,tang2020rapidly}, they are small in scale and cannot be used for training our models. We tackle the issue of data shortage by fine-tuning pre-trained biomedical language models with a small in-domain dataset. Though these models are not trained on COVID-19 data, they allow our system to warm start with general biomedical terminology. Other COVID-19-related question-answering systems have been created in recent months~\cite{Bhatia2020AWSCA,10.1145/3442381.3449991,reddy2020end}. However, our system incorporates multiple state-of-the-art information retrieval techniques with dense retrieval and BM25~\cite{robertson2009probabilistic} and the additional functionality of diversity re-ranking and multiple answer spans.

Our system is comprised of two models: the retrieval model and reading comprehension model. Our system consists of several layers of document and answer re-ranking to increase both quality and diversity in our answers. The overall system can be seen in Figure \ref{fig:overall}. We additionally provide code\footnote{\url{https://github.com/sharonlevy/Open\_Domain\_COVIDQA}} to create an online demo site to visualize our system and provide multiple filters for users to further refine their queries.

Our contributions are
\begin{enumerate}
  \item We set a precedent for quickly creating an effective open-domain question-answering system for an emergent domain.
  \item We integrate multiple stages of document re-ranking throughout our pipeline to provide relevant and diverse answers.
  \item We create an online demo to allow the public to easily obtain answers to COVID-19-related questions from credible scientific sources.
\end{enumerate}

\section{Retrieval}
The retrieval model consists of a dense retriever and contains further layers of re-ranking. In the following sections, we describe the data used to train our model, along with the model details and re-ranking strategies.
\subsection{Data}\label{sec:retdata}
As mentioned in Section \ref{sec:intro}, several COVID-19-related datasets have been published throughout the pandemic. However, there are a limited number of sizable datasets focused on the general areas of information retrieval and question-answering. In order to train on in-domain data, we utilize the COVID-QA~\cite{moller-etal-2020-covid} dataset to fine-tune our model for the document retrieval task. COVID-QA is a COVID-19 question-answering dataset and contains multiple question-answer pairs for each context document (2,019 QA pairs in total), where the documents are COVID-19-related PubMed\footnote{\url{https://pubmed.ncbi.nlm.nih.gov/}} articles. 

In order to transform the question-answering dataset for our retrieval task, we choose to utilize the questions and their related context articles during training.
We split each context article into size 100-200 tokens. Given the answer for each question and context article pair, we extract only the chunks of text that contain the answer with simple string matching and use this as a positive sample for each question. We further partition the dataset into training, development, and test sets. These splits are made at 70\%, 10\%, and 20\%, respectively. Additionally, we remove any document-specific questions (e.g. How many participants are there in this study?) from the test set for a fair assessment. 

\begin{figure}[t]
  \centering
  \includegraphics[width=\linewidth]{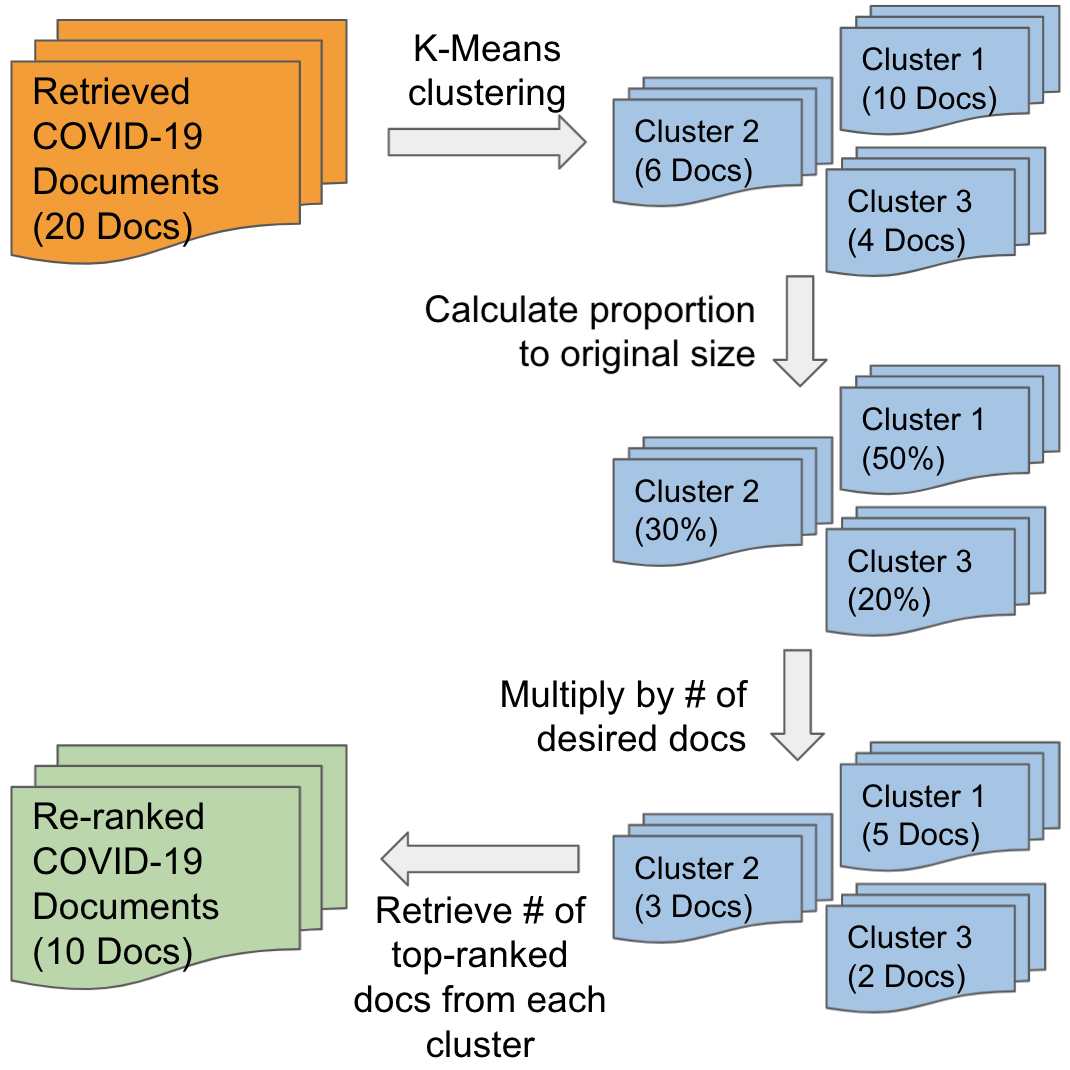}
  \caption{An outline of the diversity re-ranking process discussed in Section \ref{sec:diversity}. After the retrieval size for each cluster is calculated, the top-ranking documents (as determined by the hybrid model) are selected from each cluster according to this size and accumulated into the final set of retrieved documents. This final set is also ordered according to the original ranking by the hybrid model. }\label{fig:kmeans}
\end{figure}

We utilize the CORD-19~\cite{wang-etal-2020-cord} dataset as our document corpus for the open-domain retrieval task. The corpus website is consistently updated with newly published COVID-19-related papers from several sources. Similar to the COVID-QA dataset, we pre-process each article by splitting it into multiple document entries based on paragraph text cutoffs. Paragraphs that are longer than 200 tokens are split further until they reach the desired 100-200 token size.

\subsection{Dense Retriever}
The dense retriever consists of a unified encoder for encoding both questions and text documents. We utilize the pre-trained PubMedBERT model~\cite{gu2020domain} as the encoder and fine-tune on the COVID-QA dataset. We utilize both positive and negative samples during training. Positive samples consist of paragraphs that contain the exact answer span for the current question. Likewise, negative samples consist of paragraphs that do not contain the exact answer. 

During training, the model learns to encode questions and positive paragraphs into similar vectors such that positive paragraphs are ranked higher than negative paragraphs in similarity. After training, the CORD-19 document corpus is passed through the trained encoder and the embeddings are indexed and saved. During test time, the question is used as input to the model. The resulting embedding is used to find similarly embedded documents from the existing dense document embeddings using inner product similarity scores.

\begin{table}[t]
\centering
\begin{tabular}{l|c|c|c}
\toprule
Model &  FM@5 & FM@20 & FM@50 \\
\toprule
Dense Retrieval & 0.300 & 0.471 & 0.556\\
\hline
BM25 & 0.346 & 0.486 & 0.556\\
\hline
Hybrid Model & \textbf{0.362} & \textbf{0.498} & \textbf{0.607}\\
\bottomrule
 \end{tabular}
\caption{Comparison of dense retriever, BM25, and hybrid models for open-domain retrieval on the test set of COVID-QA. Results are evaluated with fuzzy matching (FM) scores at various retrieval count thresholds. The fuzzy matching process is described in Section \ref{sec:fm}.}\label{tab:denseresults}
\end{table}

\subsection{BM25 Re-ranking}
While the dense retriever excels in the retrieval of documents with semantic similarity to a query, there may be specific keywords in the query that are important for document retrieval. This is especially true in biomedical domains, such as COVID-19, which heavily rely on particular terminology. As a result, our system includes a second stage during retrieval in which we re-rank the top-$n$ retrieved documents with the BM25 algorithm. Specifically, we use the BM25+ algorithm defined in \cite{10.1145/2063576.2063584}. BM25 depends on keyword matching and ranks documents based on the appearance of query terms within the document corpus. We further simplify this by first removing stop words from the top-$n$ documents before re-ranking. We define the combination of our dense retriever with BM25 re-ranking as our hybrid model.

\begin{table*}[t]
\centering
\begin{tabular}{l|l|c|c}
\toprule
Model &  Datasets & Exact Match & F1  \\
\toprule
BERT & COVID-QA& 12.27& 39.07\\
\hline
BERT & SQUAD2.0& 29.24& 59.34\\
\hline
BioBERT & SQUAD2.0& 30.54 & 59.39\\
\hline
BERT & SQUAD2.0 + COVID-QA& 33.68& 65.53\\
\hline
BioBERT & SQUAD2.0 + COVID-QA& 37.59 & 66.67\\
\hline
BioBERT w/ multiple answer spans
 & SQUAD2.0 + COVID-QA & \textbf{39.16} & \textbf{72.03} \\
\bottomrule
 \end{tabular}
\caption{Comparison of BERT and BioBERT models fine-tuned on combinations of COVID-QA and SQuAD2.0. The final row includes the BioBERT model with multiple answer spans extracted. Each model was evaluated on a held-out test set from COVID-QA.}\label{tab:results}
\end{table*}

\subsection{Retrieval Diversity}\label{sec:diversity}
Following the re-ranking of retrieved documents with BM25, we aim to increase the diversity of these documents so that a user does not view nearly identical texts. To do this, we cluster the top-$k$ re-ranked documents into three clusters with K-Means clustering~\cite{macqueen1967some} and TF-IDF features. For each cluster, we compute its size in proportion to $k$. This relative size is multiplied by the desired number of documents $l$ (where $l < k$) to be retrieved. Given the resulting size for each cluster, the most relevant (top-ranked) documents are chosen in their current ranking order. This procedure is illustrated in Figure \ref{fig:kmeans}. Following this method allows us to present the user with more diverse and relevant documents that would otherwise be ranked lower.

\subsection{Retrieval Experiments}\label{sec:fm}
We use the test subset of the COVID-QA dataset to evaluate our retrieval model. However, as COVID-QA is intended for the question-answering task, we cannot accurately evaluate our model by simply calculating the retrieval rank of the correct document. This is due to our specific task of open-domain question-answering, in which we are retrieving from the large CORD-19 corpus instead of the much smaller pool of documents in COVID-QA. As a result, we define a fuzzy matching metric to evaluate the quality of our retrieved documents. This is a combination of deep semantic matching and keyword matching. We have varying combinations and thresholds based on respective conditions, such as differing answer lengths. We evaluate the answer in each QA pair in our COVID-QA test set against each retrieved document.

The deep semantic matching is achieved through the Sentence-BERT model~\cite{reimers-gurevych-2019-sentence} and F1 score is utilized for keyword matching. Each retrieved document is split into a list of sentences and each sentence is evaluated for three conditions: 
\begin{enumerate}
  \item Cosine similarity score that is greater than or equal to threshold $a$ of the sentence/query pair encoded with Sentence-BERT.
  \item Cosine similarity score greater than or equal to threshold $b$, where $b$ < $a$, and F1 score greater than or equal to threshold $c$.
  \item F1 score greater than or equal to threshold $d$, where $d$ > $c$. This is only calculated if the token count of an answer is less than or equal to 3.
\end{enumerate}
If any of the three conditions are achieved for any sentence within the retrieved document, the document is evaluated as a positive retrieval and containing the answer to the query.

We show the impact of the BM25 re-ranking stage in the hybrid model in Table \ref{tab:denseresults}. It can be seen that individually, BM25 and the dense retriever models obtain similar retrieval results. However, the hybrid model of dense retrieval followed by BM25 re-ranking allows the system to obtain more relevant documents for the user.

\section{Reading Comprehension}
The second stage of our system consists of a reading comprehension model that can answer the original query based on the retrieved documents. We describe the training data, model design, and document re-ranking associated with our model in the following sections.
\subsection{Data}
We utilize the COVID-QA dataset to train our model for the reading comprehension task. Unlike the retrieval model, the reading comprehension model utilizes both questions and answers, along with their respective context articles for training. As mentioned in Section \ref{sec:retdata}, we partition the dataset into training, development, and test sets and utilize this to evaluate the model.

\subsection{Methodology}
The reading comprehension model performs extractive question-answering. Given a question and paragraph pair, the model learns to find start and end tokens to represent the answer span (or spans) in the paragraph text. This is done by choosing the highest-ranked start and end tokens produced by the model where the start token is earlier than the end token in the text sequence. We utilize a variant of BioBERT~\cite{10.1093/bioinformatics/btz682} that is fine-tuned on the SQuAD2.0~\cite{rajpurkar-etal-2018-know} dataset\footnote{\url{https://huggingface.co/ktrapeznikov/biobert\_v1.1\_pubmed\_squad\_v2}}. We find that fine-tuning this model on COVID-QA allows the model to train on both in-domain (COVID-QA) and out-domain (SQuAD2.0) data and increases results for this task when evaluated on the test set of COVID-QA.

\begin{figure}[t]
  \centering
  \includegraphics[width=\linewidth]{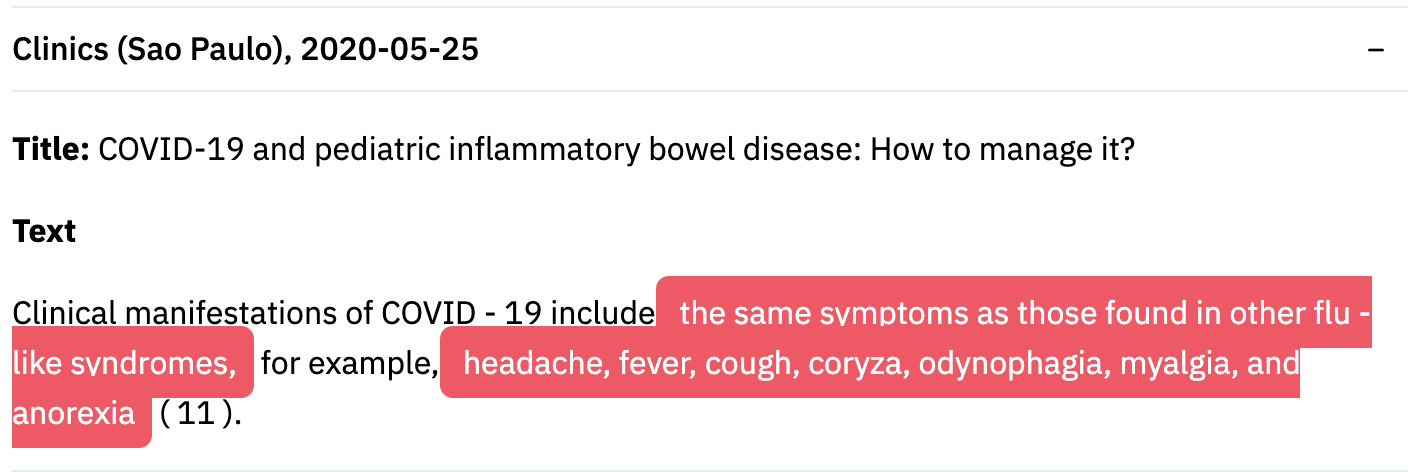}
  \caption{An example of returning multiple answers to a user for the query: ``What are symptoms of covid?''}\label{fig:multianswers}
\end{figure}

\begin{figure}[t]
  \centering
  \includegraphics[width=0.7\linewidth]{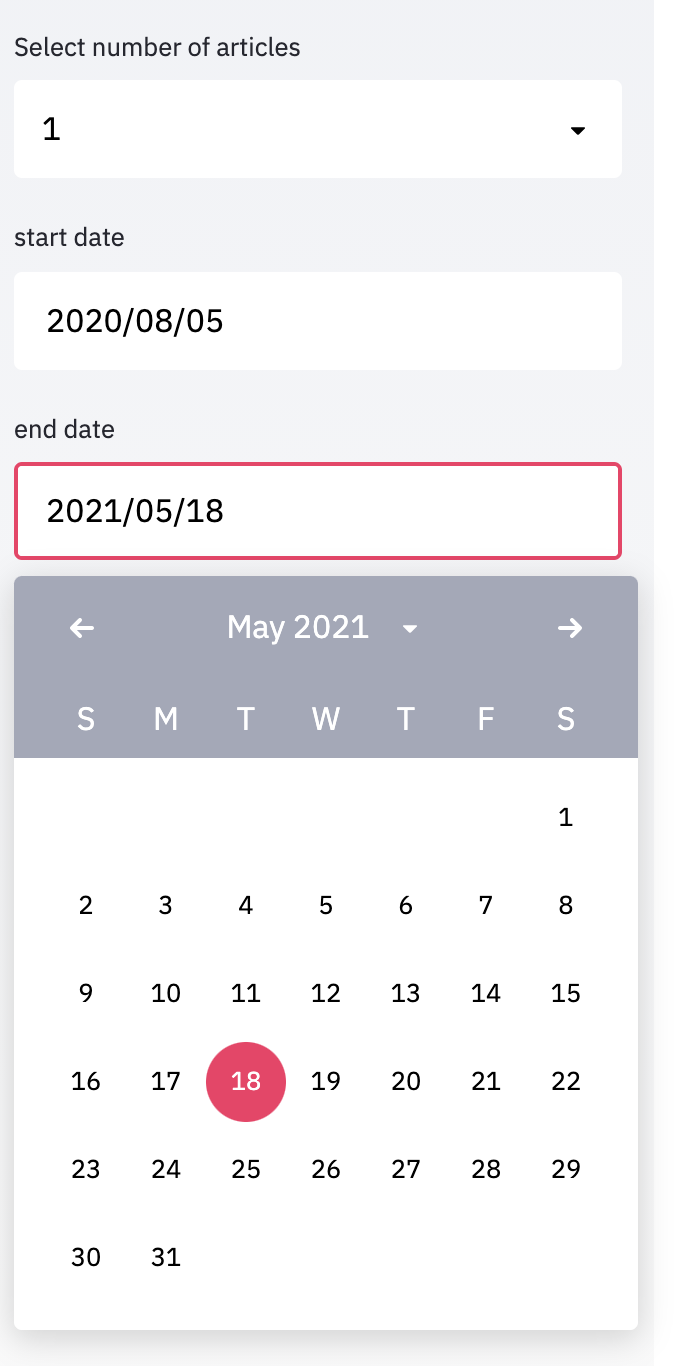}
  \caption{The side panel in the demo website which allows users to filter the number of documents retrieved and the date range for the publication date of these documents.}\label{fig:sidebar}
\end{figure}

\subsection{Multiple Answers}
Some retrieved documents may contain answer spans that are not contiguous. In order to accommodate this, we rank the top-$m$ start and end tokens according to confidence scores and select the pairs of tokens that do not overlap with higher-ranked answer spans. This allows each document to highlight up to $m$ answers rather than just one answer and increases evaluation results. We show the effect of adding multiple answer spans in Table \ref{tab:results} in comparison to various model and fine-tuning dataset combinations. An example of multiple answer spans for a given query can be seen in Figure \ref{fig:multianswers}.

\subsection{Document Re-ranking}
When the reading comprehension model is utilized in the overall system, it is used to answer the same question within a set of documents retrieved from the hybrid retriever model. While the documents are already re-ranked by the retriever, we further re-rank these documents again following the answer extraction portion of the system. When answering a question for each document, the reading comprehension model provides a confidence score alongside each start and end token. We utilize these confidence scores and reorder the current set of retrieved documents based on the combination of the start and end scores for the top answer in each document. As a result, if a question is not easily answered in a highly ranked retrieved document, the respective document will subsequently be moved to a lower rank.

\section{Open-domain Question Answering}
In the previous sections, we describe the retrieval and reading comprehension models. We combine the two models for the end-to-end open-domain question-answering task. The full system overview can be seen in Figure \ref{fig:overall}. Once the retriever is trained, the CORD-19 corpus is encoded and stored. When a user queries the system with a question, this question is encoded using the unified retriever model and the resulting vector is used to retrieve similar documents from the dense corpus. Once the top documents are retrieved, they are re-ranked with the BM25 algorithm and further clustered/re-ranked to introduce diversity to the results. The top remaining documents are used as input to the reading comprehension model along with the initial question. This model computes the answer span (and potentially spans) for each document. The documents are then re-ranked given the reading comprehension model’s confidence score in the top answer span and the answers for each document are highlighted.

\section{Demo}
We build an online demo that allows users to easily utilize our system. This website is powered through Streamlit\footnote{\url{https://streamlit.io/}}. 

\subsection{Query Filters}
 The input documents for the demo are from the CORD-19 corpus. These documents are pre-encoded by the trained hybrid retrieval model. We include several features for users to filter in order to narrow down their search. A user is able to decide how many documents they would like to be retrieved (in the range from 1 to 5) from the drop-down menu. We include start and end date selection boxes to allow users to further filter the retrieved documents by publication date within the top retrieved documents. These components are shown in Figure \ref{fig:sidebar}. If there are no documents available for the date range, we show this as a message and instead retrieve relevant documents from any date range for the user. 

\begin{figure}[t]
  \centering
  \includegraphics[width=\linewidth]{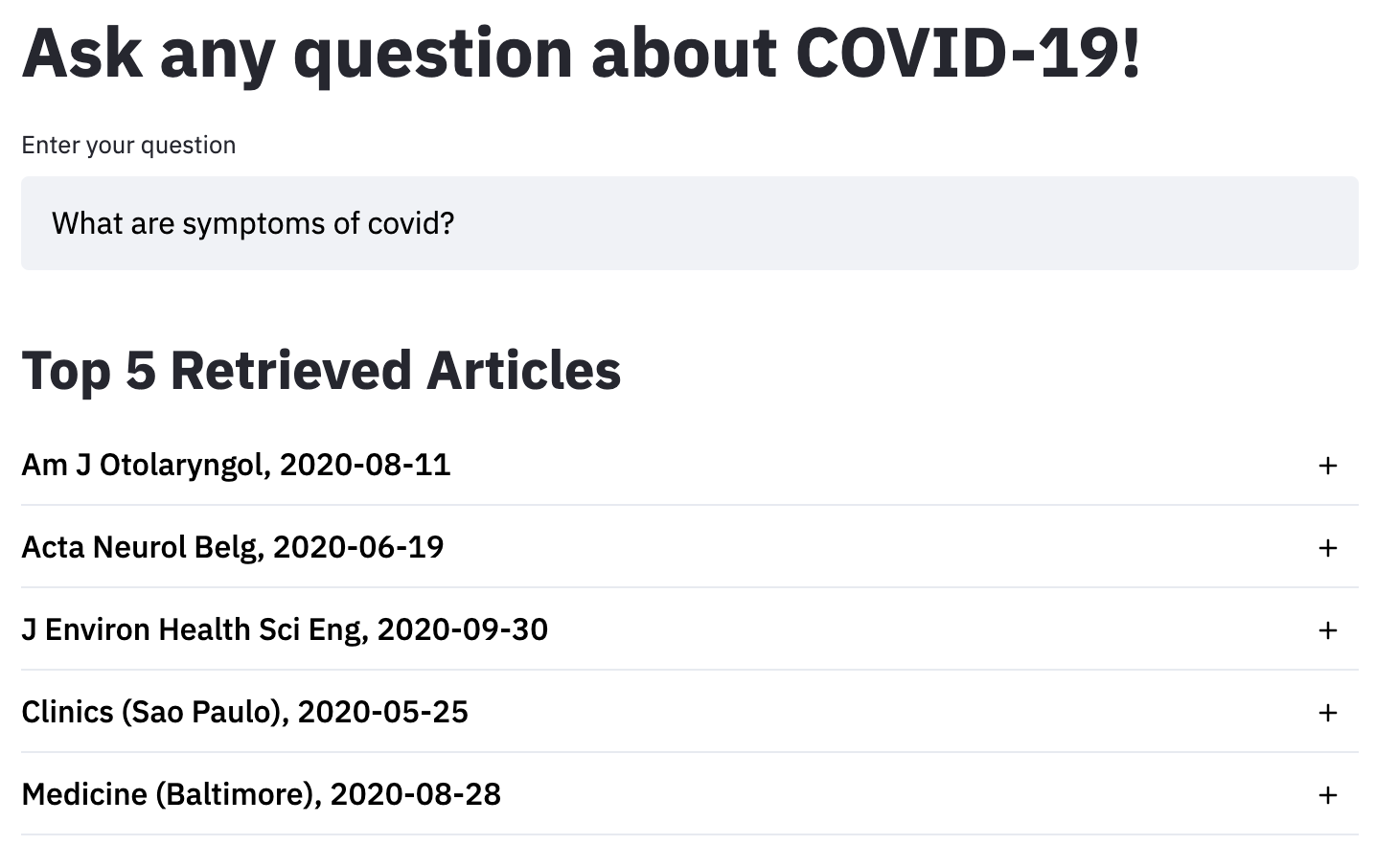}
  \caption{The list of documents returned to a user for a given query. Each document is labeled by its publishing journal and publication date.}\label{fig:docs}
\end{figure}

\begin{figure}[t]
  \centering
  \includegraphics[width=\linewidth]{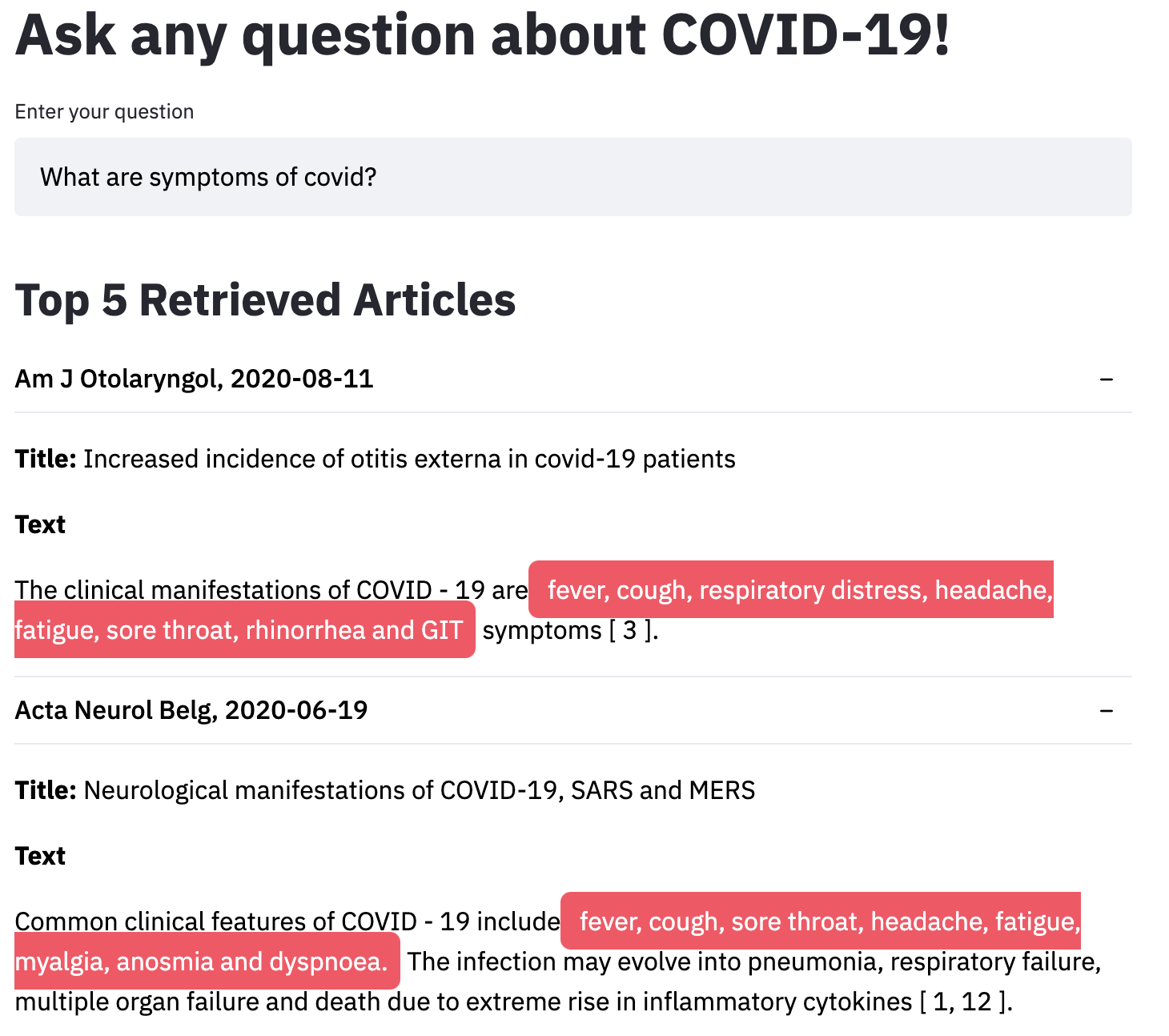}
  \caption{Retrieved documents for a given query can be expanded to show their respective article titles and text snippets. Extracted answers for each document are highlighted in red.}\label{fig:answers}
\end{figure}

\subsection{Demo Procedure}
The user can enter a free-text question in English into the search bar as seen in Figure \ref{fig:docs}. This question is encoded by the trained retrieval model and used to find matching documents. The reading comprehension model uses the retrieved documents and query to extract the answer (or answers) and re-rank the documents based on the answer confidence scores. The chosen number of retrieved documents is displayed to the user. Each document is displayed alongside its journal or source name and publication date from its respective CORD-19 article. The user can expand each document heading to view the article title and text snippet. The extracted answers are highlighted in red as seen in Figure \ref{fig:answers}. 

\section{Conclusion}
In this paper, we present an open-domain question answering system for the emergent domain of COVID-19. Our system is comprised of retrieval and reading comprehension components, with several layers of refinement to increase the quality and diversity of responses. The system allows users to quickly search COVID-19-related questions and obtain a set of answers from biomedical publications. Additionally, we provide a demo website that allows users to easily interact with our system and apply additional filters to further refine their search. We hope that amidst the time of a global pandemic, our system can serve as both a resource in finding credible answers to users' COVID-19 questions and a model for future systems in similar emergent domains.  

\section*{Acknowledgements}
This work is partly sponsored by Office of the Director of National Intelligence/Intelligence Advanced Research Projects Activity (IARPA). The views and conclusions contained in this document are those of the authors and should not be interpreted as representing the official policies, either expressed or implied, of the U.S. Government. The U.S. Government is authorized to reproduce and distribute reprints for Government purposes notwithstanding any copyright notation herein.

\bibliography{anthology,custom}
\bibliographystyle{acl_natbib}


\end{document}